# Swarming around Shellfish Larvae

Vitorino Ramos[1], Jonathan Campbell[2],
John Slater[3], John Gillespie[3], Ivan F. Bendezu[3], Fionn Murtagh[4]

[1]CVRM-IST, Technical University of Lisbon, Av. Rovisco Pais, 1049-001, Lisbon, PORTUGAL
[2]Department of Computing, Letterkenny Institute of Technology,
Port Road, Co. Donegal, IRELAND (*jonathan.campbell@lyit.ie*)
[3]Department of Science, Letterkenny Institute of Technology, Port Road, Co. Donegal, IRELAND
[4]Dept. of Computer Science, Royal Holloway, Univ. of London, Surrey, UNITED KINGDOM

**Abstract:** The collection of wild larvae seed as a source of raw material is a major sub industry of shellfish aquaculture. To predict when, where and in what quantities wild seed will be available, it is necessary to track the appearance and growth of planktonic larvae. One of the most difficult groups to identify, particularly at the species level are the Bivalvia. This difficulty arises from the fact that fundamentally all bivalve larvae have a similar shape and colour. Identification based on gross morphological appearance is limited by the time-consuming nature of the microscopic examination and by the limited availability of expertise in this field. Molecular and immunological methods are also being studied. We describe the application of computational pattern recognition methods to the automated identification and size analysis of scallop larvae. For identification, the shape features used are binary invariant moments; that is, the features are invariant to shift (position within the image), scale (induced either by growth or differential image magnification) and rotation. Images of a sample of scallop and non-scallop larvae covering a range of maturities have been analysed. In order to overcome the automatic identification, as well as to allow the system to receive new unknown samples at any moment, a self-organized and unsupervised ant-like clustering algorithm based on Swarm Intelligence is proposed, followed by simple *k*-NNR nearest neighbour classification on the final map. Results achieve a full recognition rate of 100% under several situations ($k$ =1 or 3).

**1. Background and Objective**

Diminishing fishery stocks increased fishing pressure, and the ever present threat of pollution mean that the seas can no longer be regarded as a limitless food source. Aquaculture production is expanding rapidly as the way forward for the marine food sector to develop a sustainable industry for present and future generations. Production valued in excess of 114 million € (Euro) occurred in 2002 and is projected to increase to 97,023 tonnes valued at over 175 million € by 2008. Currently approximately 60% of aquaculture volume comprises shellfish with much of this production, dependent on the collection of wild seed as a source of raw material. To predict when, where and in what quantities wild seed will be available, knowledge of the adult shellfish spawning pattern and the dispersal of planktonic bivalve larvae in the waters around our European coast are required. Whilst large numbers of plankton samples are relatively easy to collect and good taxonomic keys exist for many of the specimens in such samples, one of the most difficult groups to identify, particularly at the species level are the Bivalvia. This difficulty arises from the fact that fundamentally all bivalve larvae have a similar shape, they are all brownish in colour to one degree or another, and there are no protruding parts or appendages, which can be used to aid identification [3,4,6]. Molecular methods based on antibody and oligonucleotide markers have been developed by *Andre*, *Paugham*, *Frischer*, and *Hare* et al. The application of an immunological method of larvae identification has recently been reported by *Paugham* et al. A useful survey covering the full spectrum of methods (as those above) is given in *Garland* and *Zimmer* [15]. The paper clearly identifies the labour intensity of the problem, especially considering the large number of samples that could be generated by the necessary field surveys. The candidate methods surveyed are: (a) morphological (based on microscopy); (b) optical, e.g. fluoroscopy, but it is concluded that this is unlikely to provide *species* discrimination, only *class* discrimi-

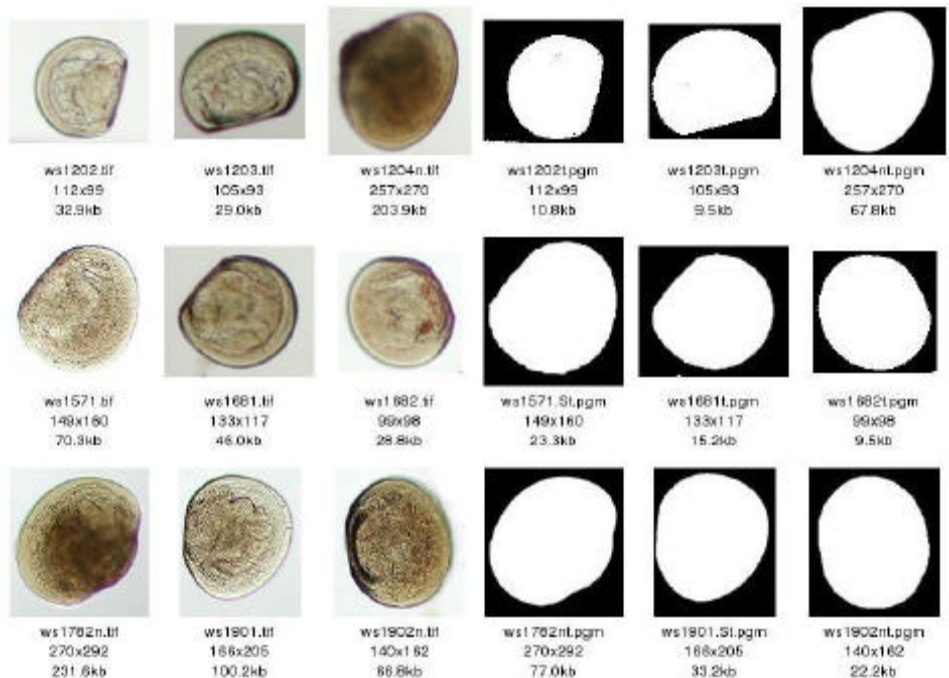

Fig. 1 – On the left a compendium of 9 raw images (out of 20 samples) used in the present study. Samples 7 (2nd row, 1st column) and 14 (3rd row, 2nd column) are scallops. Respective segmented images on the right (.S in the name indicates scallop samples).

-nation; (c) molecular (immunological and DNA-based). In the context of (a) (morphological), scanning electron microscope (SEM) and optical microscope methods are mentioned. Other methods are described in [16] for automated identification of several species (not only Bivalvia). The use of simple shape features based on area and bounding rectangle dimensions is reported in *de Pontual* et al [33]; however, this study seems to have been limited by the availability of methods offered by the image processing package used, and the article admits difficulties in handling the wide range of species that may be encountered. Complex texture-based image processing of polarization microscopy images is described in *Tiwari* and *Gallager* [42], i.e. in contrast to [33] which considers only the shape outline/silhouette, this paper considers the texture (colour patterns) that show up in the (polarized) images.

## 2. Methods

A chief aim of the methods described here is to avoid the need for exotic and expensive sensing, i.e. to perform automatic image analysis of plain optical microscopy images. We start with a raw image such as that shown above; of course, a microscope image of a plankton sample is much larger and contains a great many larvae, perhaps representing a range of species, and other matter. The first stage depends on what is to follow but a typical requirement is to *segment*, i.e. separate the object from the background. There are many candidate segmentation algorithms [32], many of them based on statistical clustering. Above we show the result of segmenting using a simple threshold technique in which the threshold is based on the image histogram [18]. This segmentation is relatively trivial. However, in other cases, the translucency of the objects is more pronounced, i.e. some parts of the object are actually less optically dense than the background, thereby defeating simple thresholding. We are currently studying methods based on Markov random field models [40,30], which combine spatial contiguity with colour/grey-level similarity. In the segmented image, each object is reduced to a planar shape from which shape features can be extracted. Since objects can appear anywhere, and at any orientation, we require *shift-* and *rotation-invariant* features. Invariant moments [20,31,28], are ideal candidates as described in [7,3]. Invariant moments (of a binarised object), ignore object texture, i.e. in contrast to [42] discussed above. We believe that this is well founded, since manual inspection suggests that discrimination should be possible based on outline shape alone. In our present work, following feature extraction, feature vectors are passed to a classification algorithm based on a self-organized clustering [36] and *k*-NNR [10,13]. Section 3 is devoted to introduce these invariant moment features, while section 4 describes our algorithm, and results achieved over several scallop and non-scallop shellfish larvae image samples are presented in section 5.

**3. Invariant Moment features**

Invariant moments were earlier introduced by [20]. Further elaborations and discussions of computer implementations are discussed in [28,7,6]. What we require are features that capture the shape of objects. Since an object can appear in any position in the image, we require *shift invariant* features; likewise, since an object must be recognised independently of its angular orientation, we require *rotationally invariant* features. Finally, because we want to remove image magnification and other size effects, we require *scale* (or size) invariance. When applied to objects which have been segmented into *object* and *background* – $f(r,c) \in {0,1}$, where 1 corresponds to object and 0 corresponds to background, moments and moment invariants capture the shape of the object silhouette. For discrete images the two-dimensional moment of order $p,q$ is defined as Eq. 1, where $f(r,c)$ is the image value at row $r$ and column $c$. In the case of binary images, $f(r,c) \in {0,1}$, so that $m_{00} = \Sigma r \Sigma c \, r^0 c^0 f(r,c) = \Sigma r \Sigma c \, f(r,c)$, corresponds to area measured in pixels. Considering pixels as point masses, the centre of mass of the image (i.e. of the segmented object) is given by Eq. 2. It is easy to achieve *shift invariance* by referring all moment calculations to the centre of mass given by Eq. 2; these are the so-called centralised moments (Eq. 3). Then, *scale invariance* is achieved by normalising with respect to area ($m_{00}$), i.e., if we (notionally) scale the row and column dimensions by Eq. 4. We then arrive at *normalised central moments* (Eq. 5) where $g = 1+[(p+q)/2]$, being $p+q \geq 2$. The later equation is derived as follows. Consider a continuous image (rather that one with discrete pixels) described by Eq. 6. Now scale each dimension by $\lambda$. (Eq. 7), leading to Eq. 8.

$$m_{pq} = \sum_r \sum_c r^p c^q f(r,c) \qquad (1) \qquad \bar{r} = \frac{m_{10}}{m_{00}}, \bar{c} = \frac{m_{01}}{m_{00}} \qquad (2)$$

$$m'_{pq} = \sum_r \sum_c (r-\bar{r})^p (c-\bar{c})^q f(r,c) \qquad (3) \qquad \lambda = \frac{1}{\sqrt{m_{00}}} \qquad (4)$$

$$n_{pq} = \frac{m'_{pq}}{\sqrt{m_{00}}^\gamma} \qquad (5) \qquad m_{pq} = \int_{-\infty}^{\infty} \int_{-\infty}^{\infty} r^p c^q f(r,c) dr dc \qquad (6)$$

$$m_{\lambda,pq} = \int_{-\infty}^{\infty} \int_{-\infty}^{\infty} r^p c^q f(r/\lambda, c/\lambda) dr dc \quad (7) \qquad m_{\lambda,pq} = \lambda^{2+p+q} m_{pq} \qquad (8)$$

$$m'_{pq} = \sum_r \sum_c (r-\bar{r})^p (c-\bar{c})^q f(r,c) \qquad (9) \qquad m'_x = \sum_x (x-\bar{x})^p f(x) \qquad (10)$$

**3.1. Interpretation of Moments**

Here we give a brief interpretation of (*centralised*) moments, i.e. moments which are computed with respect to the centre of mass Eq. 3, has described by Eq. 9. Let us first examine one-dimensional moments - which are often used in describing one-dimensional probability mass functions, has Eq. 10. Here, $m'_0$ is the sum of $f(x)$, $m'_1$ is the mean of $f(x)$ about $x?$, i.e. 0, since $m'_1 = x?$. On the other hand, $m'_2$ is computed by weighting $f(x)$ by $(x-x?)^2$, so $m'_2$ is a measure of the "fatness" of $f(x)$ ; in probability mass functions it is the *variance*. Now $m'_3$ is computed by weighting $f(x)$ by $(x-x?)^3$; note that by weighting $f(x)$ by $(x-x?)^3$ pixels on the negative side of $(x-x?)$ contribute negatively, whilst pixels on the positive side of $(x-x?)$ contribute positively, i.e. $m'_3$ measures *skewness* (lack of symmetry) about $x?$. If $f(x)$ has a long tail to the right (increasing $x$), positive skewness, $m'_3 > 0$, will result; correspondingly, a long tail to the left will yield $m'_3 < 0$; and $f(x)$ symmetric about the mean $x?$ will have $m'_3 = 0$.

$$\begin{aligned}
h_1 &= n_{20} + n_{02}, \\
h_2 &= (n_{20} - n_{02})^2 + 4n_{11}^2, \\
h_3 &= (n_{30} - 3n12)^2 + (3n_{21} - n_{03})^2, \\
h_4 &= (n_{30} + n_{12})^2 + (n_{21} + n_{03})^2, \\
h_5 &= (n_{30} + 3n12)(n_{30} + n_{12}) \\
&\quad + ((n_{30} + n_{12})^2 - 3(n_{21} - n_{03}^2)), \\
&\quad + (3n_{21} - n_{03}) \\
&\quad (n_{21} + n_{03})(3(n_{30} + n_{12})^2 - (n_{21} + n_{03})^2), \\
h_6 &= (n_{20} - n_{02})((n_{30} + n_{12})^2 - (n_{21} + n_{03})^2) \\
&\quad + 4n_{11}(n_{30} + n_{12})(n_{21} + n_{03}), \\
h_7 &= (3n_{21} - n_{03})((n_{30} + n_{12})(((n_{30} + n_{12})^2 \\
&\quad - 3(n_{21} + n_{03})^2 \\
&\quad (3n_{21} - n_{03})(n_{21} + n_{03})(3(n_{12} + n_{30})^2 - (n_{21} + n_{03})^2)
\end{aligned}$$

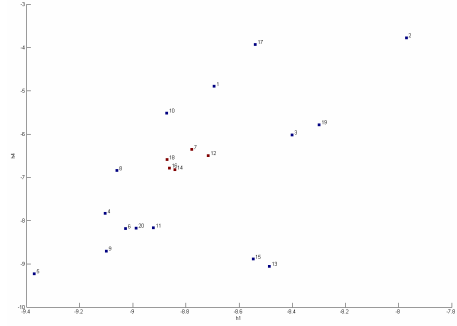

Fig. 2 – *h1* x *h4* invariant moments scatter plot. Five scallop samples (7, 12, 14, 16 and 18) appear together in the middle region.

Let us return now to two dimensions and Eq. 9. Here, $m'_{00}$ is the sum of $f(r,c)$, i.e. in our binary image case, it is the total area of object pixels (=1); $m'_{10}$ is the *r*-centre of mass; $m'_{10} = 0$ since $m'_{pq}$ are referred to the centre of mass. Likewise $m'_{01} = 0$ is the centre of mass; $m'_{20}$ measures the "fatness" (variance in statistic terms) in the *r*-dimension. Likewise $m'_{02}$ is the fatness *c*-dimension. For example, long thin ellipse aligned vertically along the *r*-axis will have a large $m'_{20}$ and small $m'_{02}$, with a long thin ellipse aligned horizontally will have large $m'_{02}$ and small $m'_{20}$; $m'_{30}$ measures the skewness in the *r*-dimension and $m'_{03}$ measures the skewness in the *c*-dimension. An "egg" shape aligned vertically will have a relatively large $m'_{30}$; likewise the same shape aligned horizontally will have a relatively large $m'_{03}$. Axially symmetric shapes, for example ellipses (and circles) have $m'_{03} = m'_{30} = 0$; $m'_{11}$ measures the degree to which the object is aligned along the *r-c*-diagonals; its sign indicates the proportion contributed by the four diagonals. Shapes such as ellipses aligned along one or other axis have $m'_{11} = 0$. Finally $m'_{21}$ and $m'_{12}$ provide measures of skew-symmetry, i.e. symmetry along a horizontal. However, the problem of *rotational invariance* remains. One solution, as in [28], computes the angles of the principal axes and then recomputes $m'_{\lambda,pq}$ such that moments are referred to these axes (the term *principal axes* is used to convey the same meaning as "principal axes of an ellipse"). Unfortunately, when principal axes are indistinct, e.g. as in a disc (circular) shape, this normalisation can sometimes fail. Hence the interest in Hu's [20] rotationally invariant moments. Under scale normalisation and shift invariance, moments $m_{00}$ (area) may be discarded, and location moments: $m_{10}$ and $m_{01}$ are both zero. Hu's rotationally invariant moments are derived from the normalised central moments of Eq. 5. Then we arrive at invariant moments *h1*, *h2*,…, *h7* which are subjected to further normalisation transformations, mostly involving logarithms [28] to yield values which are of broadly comparable size and range. When, following this order, the first seven invariant moments [28] are extracted and normalised, we finally arrive at those values accessible in table 1.

### 3.2. Mapping Invariant Moments

Plotting some combinations of these feature set can yield interesting results. For instance the *h1* x *h4* scatter plot (fig.2) seems to help discriminate scallop samples. Indeed, samples 7,12,14,16 and 18 appear together in the middle of the plotting area, far from the remaining non-scallop samples. However, it is hard to say if this always happens when new samples arrive to the system in order to be classified (for instance, using *k*-NNR nearest neighbour rule [10,13] after mapping some successful combinations of these invariant moment features). In order to reduce the number of features or pinpoint those that are relevant, more robust and analytical methods should be used, as Principal Component Analysis (PCA, [22]). Other possibility, is not only to try to reduce the number of features as at the same time, trying to increase the recognition rate, which conducts us to a multi-modal optimization problem. A successful method using Genetic Algorithms (Evolutionary Computation) was designed earlier by *Ramos* and *Pina* [39], in order to classify 187 images (each described by 117 Mathematical Morphology [24] features) from 14 classical types of Portuguese granites. Their method not only increased the recognition rate to 100% as features were reduced from 117 to 3, representing simultaneously a robust strategy in order to understand the proper nature of the images treated, and their discriminant features. In the present case, however, the numbers of samples as well as the number of features are not elevated, and instead, a self-organized ant-like clustering method followed by *k*-NNR classification is proposed, since it can be scaled to a large range of conditions (number of samples, number of features, or eventually and simultaneously treat samples never yet processed).

| Image | h1 | h2 | h3 | h4 | h5 | h6 | h7 |
|---|---|---|---|---|---|---|---|
| 1 | -8,6940 | -7,9026 | -12,2217 | -4,9005 | 0,1141 | -0,198 | 0,0804 |
| 2 | -7,9710 | -5,4640 | -11,8688 | -3,7754 | -0,2349 | -0,7533 | -0,0871 |
| 3 | -8,4007 | -6,8575 | -11,5683 | -6,0194 | -0,0865 | -0,3836 | 0,0658 |
| 4 | -9,1047 | -10,4998 | -12,4660 | -7,8341 | -0,0486 | -0,2176 | 0,0245 |
| 5 | -9,3712 | -13,8080 | -12,8727 | -9,2301 | -0,0002 | -0,0028 | 0,0000 |
| 6 | -9,0280 | -9,5743 | -12,4695 | -8,1891 | 0,0332 | -0,0844 | 0,0218 |
| *7 | -8,7786 | -8,2680 | -12,0012 | -6,3576 | -0,0683 | -0,1458 | -0,0551 |
| 8 | -9,0596 | -10,4306 | -12,3343 | -6,8460 | -0,0597 | -0,2134 | 0,0425 |
| 9 | -9,1003 | -9,2379 | -12,8374 | -8,7028 | 0,0197 | 0,1318 | 0,0068 |
| 10 | -8,8725 | -9,5835 | -12,0148 | -5,5173 | -0,1274 | -0,3880 | 0,0646 |
| 11 | -8,9225 | -8,3671 | -12,7163 | -8,1694 | -0,0385 | -0,2324 | -0,0121 |
| *12 | -8,7167 | -7,6808 | -12,1323 | -6,4967 | 0,1003 | 0,3857 | 0,0381 |
| 13 | -8,4861 | -6,3422 | -12,8731 | -9,0545 | -0,0051 | -0,0873 | 0,0034 |
| *14 | -8,8416 | -8,2991 | -12,1866 | -6,8248 | 0,0825 | 0,3249 | 0,0398 |
| 15 | -8,5474 | -6,4951 | -12,8462 | -8,8891 | -0,0053 | 0,0707 | -0,0075 |
| *16 | -8,8622 | -9,1319 | -11,6367 | -6,7841 | 0,1023 | 0,3345 | 0,0218 |
| 17 | -8,5396 | -7,8825 | -11,4335 | -3,9287 | -0,2168 | -0,5815 | 0,1113 |
| *18 | -8,8719 | -8,6952 | -11,9684 | -6,5917 | 0,1032 | 0,3574 | -0,0280 |
| 19 | -8,2990 | -6,0410 | -12,2315 | -5,7875 | 0,1053 | 0,3277 | 0,0590 |
| 20 | -8,9878 | -8,9661 | -12,7607 | -8,1713 | -0,0348 | -0,2139 | -0,0149 |

Table 1 – Invariant Moment features (* signifies scallop).

**4. Bio-inspired self-organized ant-like clustering**

Data clustering is precisely one of those problems in which real ants can suggest very interesting heuristics for computer scientists. One of the first studies using the metaphor of ant colonies related to the above clustering domain is due to *Deneubourg* [11], where a population of ant-like agents randomly moving onto a 2D grid are allowed to move basic objects so as to cluster them. This method was then further generalized by *Lumer* and *Faieta* [25] (here after *LF* algorithm), applying it to exploratory data analysis, for the first time. In 1995, the two authors were then beyond the simple example, and applied their algorithm to interactive exploratory database analysis, where a human observer can probe the contents of each represented point (sample, image, item) and alter the characteristics of the clusters. They showed that their model provides a way of exploring complex information spaces, such as document or relational databases, because it allows information access based on exploration from various perspectives. However, this last work entitled "Exploratory Database Analysis via Self-Organization", according to [14], was never published due to commercial applications. They applied the algorithm to a database containing the "profiles" of 1650 bank customers. Attributes of the profiles included marital status, gender, residential status, age, a list of banking services used by the customer, etc. Given the variety of attributes, some of them qualitative and others quantitative, they had to define several dissimilarity measures for the different classes of attributes, and to combine them into a global dissimilarity measure (in, pp. 163, Chapter 4 [14]). More recently, *Ramos* et al. [37,36,38] presented a novel strategy (ACLUSTER) to tackle unsupervised clustering as well as data retrieval problems, avoiding not only short-term memory based strategies, as well as the use of several artificial ant types (using different speeds), present in those approaches proposed initially by *Lumer* [25]. Other works in this area include those from *Monmarché* et al. [29], *Ramos*, *Merelo* et al. [38,36,37,34], *Handl* and *Dorigo* [19], *Ramos* and *Abraham* [35,1].

## 4.1 Distributed, collaborative and Stigmergic clustering

The swarm intelligence algorithm fully uses agents that stochastically move around the classification "habitat" following pheromone concentrations. That is, instead of trying to solve some disparities in the basic LF algorithm by adding different ant casts, short-term memories and behavioral switches, which are computationally intensive, representing simultaneously a potential and difficult complex parameter tuning, it was our intention to follow real ant-like behaviors as possible (some other features will be incorporated, as the use of different response thresholds to task-associated stimulus intensities, discussed later). In that sense, bio-inspired spatial transition probabilities are incorporated into the system, avoiding randomly moving agents, which tend the distributed algorithm to explore regions manifestly without interest (e.g., regions without any type of object clusters), being generally, this type of exploration, counterproductive and time consuming. Since this type of transition probabilities depend on the spatial distribution of pheromone across the environment, the behavior reproduced is also a stigmergic one [37,11]. Moreover, the strategy not only allows to guide ants to find clusters of objects in an adaptive way (if, by any reason, one cluster disappears, pheromone tends to evaporate on that location), as the use of embodied short-term memories is avoided (since this transition probabilities tends also to increase pheromone in specific locations, where more objects are present). As we shall see, the distribution of the pheromone represents the memory of the recent history of the swarm, and in a sense it contains information which the individual ants are unable to hold or transmit. There is no direct communication between the organisms but a type of indirect communication through the pheromonal field.

In fact, ants are not allowed to have any memory and the individual's spatial knowledge is restricted to local information about the whole colony pheromone density. In order to design this behavior, one simple model was adopted (*Chialvo* and *Millonas*, [9]), and extended (as in [38,36]) due to specific constraints of the present proposal. As described in [9], the state of an individual ant can be expressed by its position $r$, and orientation $q$. It is then sufficient to specify a transition probability from one place and orientation $(r,q)$ to the next $(r^*,q^*)$ an instant later. The response function can effectively be translated into a two-parameter transition rule between the cells by use of a pheromone weighting function (Eq. 11). This equation measures the relative probabilities of moving to a cite $r$ (in our context, to a grid location) with pheromone density $s(r)$. The parameter $b$ is associated with the osmotropotaxic sensitivity (a kind of instantaneous pheromonal gradient following), and on the other hand, $1/d$ is the sensory capacity, which describes the fact that each ant's ability to sense pheromone decreases somewhat at high concentrations. In addition to the former equation, there is a weighting factor $w(Dq)$, where $Dq$ is the change in direction at each time step, i.e. measures the magnitude of the difference in orientation. As an additional condition, each individual leaves a constant amount $h$ of pheromone at the cell in which it is located at every time step $t$.

$$W(s) = \left(1 + \frac{s}{1+ds}\right)^b \quad (11) \qquad P_{ik} = \frac{W(s_i)w(\Delta_i)}{\sum_{j/k} W(s_j)w(\Delta_j)} \quad (12)$$

This pheromone decays at each time step at a rate $k$. Then, the normalised transition probabilities on the lattice to go from cell $k$ to cell $i$ are given by $P_{ik}$ [9] (Eq. 12), where the notation $j/k$ indicates the sum over all the pixels $j$ which are in the local neighborhood of $k$. $D_i$ measures the magnitude of the difference in orientation for the previous direction at time $t$-1.

## 4.2 Picking and Dropping Data-Objects

In order to model the behavior of ants associated to different tasks, as dropping and picking up objects, we suggested [36] the use of combinations of different response thresholds. As we have seen before, there are two major factors that should influence any local action taken by the ant-like agent: the number of objects in his neighborhood, and their similarity (including the hypothetical object carried by one ant). *Lumer* and *Faieta* [25], use an average similarity, mixing distances between objects with their number, incorporating it simultaneously into a response threshold function. Instead, we recommend the use of combinations of two independent response threshold functions, each associated with a different environmental factor (or, stimuli intensity), that is, the number of objects in the area, and their similarity.

**Algorithm.** High-level description of *ACLUSTER*.

```
/* Initialization */
For every object or data-item o_i do
Place o_i randomly on grid
End For
For all agents do
Place agent at randomly selected site
End For

/* Main loop */
For t = 1 to t_max do
For all agents do
sum = 0
Count the number of items n around site r

If ((agent unladen) & (site r occupied by item o_i)) then
For all sites around r with items present do
/* According to Eqs. 14, 16, 17 and 18 */
Compute d,χ, e and P_p
Draw a random real number R between 0 and 1
If (R ≤ P_p) then
sum = sum + 1
End If
End For
If ((sum = n/2) or (n = 0)) then
Pick up item o_i
End If
Else If ((agent carrying item o_i) & (site r empty)) then
For all sites around r with items present do
/* According to Eqs. 14, 15, 17 and 18 */
Compute d,χ, d and P_d
Draw random real number R between 0 and 1
If (R ≤ P_d) then
sum = sum + 1
End If
End For
If (sum = n/2) then
Drop item o_i
End If
End If

/* According to Eqs. 11 and 12 (section 4.1) */
Compute W(s) and P_ik
Move to a selected r not occupied by other agent
Count the number of items n around that new site r
Increase pheromone at site r according to n, that is:
P_r = P_r + [h + (n/a)]
End For
Evaporate pheromone by K, at all grid sites
End For
Print location of items

/* Values of parameters used in experiments */
k_1 = 0.1, k_2 = 0.3, K = 0.015, h = 0.07, a = 400,
b = 3.5, ?=0.2, t_max = 10^6 steps.
```

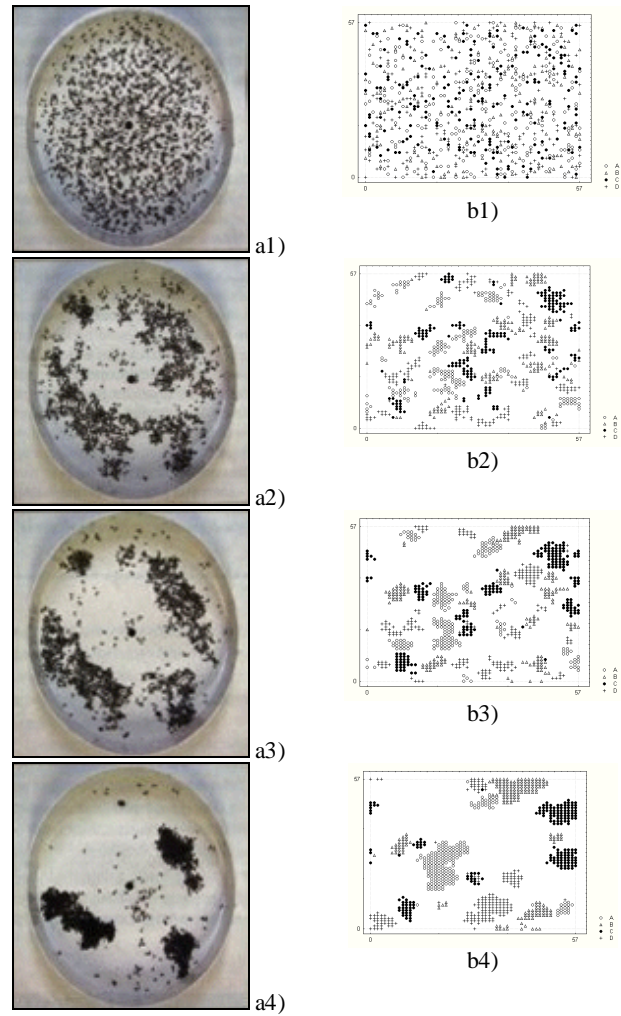

a1)

a2)

a3)

a4)

b1)

b2)

b3)

b4)

Fig. 3 – From a1) to a4), a sequential clustering task of corpses performed by a real ant colony. 1500 corpses are randomly located in a circular arena with radius = 25 cm, where *Messor Sancta* ant workers are present. The figure shows the initial state (a1), 2 hours (a2), 6 hours (a3) and 26 hours (a4) after the beginning of the experiment [14]. In b1-b4, some experiments with the present algorithm, conducted on synthetic data (as in [36,25]). Spatial distribution of 800 items on a 57 x 57 non-parametric toroidal grid at several time steps. At $t=1$, four types of items are randomly allocated into the grid. As time evolves, several homogenous clusters emerge due to the ant colony action, and as expected the global entropy decreases [37].

Moreover, the computation of average similarities are avoided in the present algorithm, since this strategy can be somehow blind to the number of objects present in one specific neighborhood. In fact, in *Lumer* and *Faieta*'s work [25], there is an hypothetical chance of having the same average similarity value, respectively having one or, more objects present in that region. But, experimental evidences and observation in some types of ant colonies can provide us with a different answer. After *Wilson* (*The Insect Societies*, Cambridge Press, 1971), it is known that minors and majors in the polymorphic species of ants *Genus Pheidole*, have different response thresholds to task-associated stimulus intensities (i.e., division of labor). Recently, and inspired by this experimental evidence, *Bonabeau* et al. [14], proposed a family of response threshold functions in order to model this behavior. According to it, every individual has a response threshold *q* for every task. Individuals engage in task performance when the level of the task-associated stimuli *s*, exceeds their thresholds. Author's defined *s* as the intensity of a stimulus associated with a particular task, i.e. *s* can be a number of encounters, a chemical concentration, or any quantitative cue sensed by individuals. One family of response functions $T_q(s)$ (the probability of performing the task as a function of stimulus intensity *s*), that satisfy this requirement is given by (Eq. 13) [14], where *n*>1 determines the steepness of the threshold (normally *n*=2, but similar results can be obtained with other values of *n*>1). Now, at *s* = *q* , this probability is exactly ½. Therefore, individuals with a lower value of *q* are likely to respond to a lower level of stimulus. In order to take account on the number of objects present in one neighborhood, Eq. 13, was used (where, *n* now stands for the number of objects present in one neighborhood, and *q* = 5), defining χ (Eq. 14) as the response threshold associated to the number of items present in a 3 x 3 region around *r* (one specific grid location). Now, in order to take account on the hypothetical similarity between objects, and in each ant action due to this factor, a Euclidean normalized distance *d* is computed within all the pairs of objects present in that 3 x 3 region around *r*. Being *a* and *b*, a pair of objects, and $f_a(i)$, $f_b(i)$ their respective feature vectors (being each object defined by *F* features), then $d = (1/d_{max}).[(1/F).\sum_{i=1,F}(f_a(i)-f_b(i))^2]^{1/2}$. Clearly, this distance *d* reaches its maximum (=1, since *d* is normalized by $d_{max}$) when two objects are maximally different, and *d*=0 when they are equally defined by the same *F* features. Moreover, *d* and *e* (Eqs. 15,16), are respectively defined as the response threshold functions associated to the similarity of objects, in case of dropping an object (Eq. 15), and picking it up (Eq. 16), at site *r*. Finally, in every action taken by an agent, and in order to deal, and represent different stimulus intensities (number of items and their similarity), present at each site in the environment visited by one ant, the strategy used a composition of the above defined response threshold functions (Eqs. 14, 15 and 16). Several composed probabilities were analyzed [36] and used as test functions in one preliminary test. The best results were achieved with the test function #1 below (Eqs. 17, 18), achieving a high classification rate (out of 4 different functions were used, as well the *LF* algorithm [25]; for comparison reasons – see [37,36]). Alternatively, the system can also be robust feeding the data continuously (for instance, as they appear) as proved in past works [35]. For other algorithm details please consult [36,37,38].

$$T_q(s) = \frac{s^n}{s^n + q^n} \quad (13) \qquad c = \frac{n^2}{n^2 + 5^2} \quad (14)$$

$$d = \left(\frac{k_1}{k_1 + d}\right)^2 \quad (15) \qquad e = \left(\frac{d}{k_2 + d}\right)^2 \quad (16)$$

$$P_p = (1-c).e \quad (17) \qquad P_d = c.d \quad (18)$$

**5. Data and results**

After invariant moment feature extraction (section 3), those values presented in table 1 were subjected to normalisation. This normalisation is based on the computation of *max* and *min* values for each of the seven invariant moments. Then, each invariant moment values (over 20 samples) are normalised to the interval [0,1], in order to be randomly introduced in the algorithm described above. In our case, each object (each shellfish larvae image sample) manipulated by the artificial ant colony is represent by a feature vector composed of 7 elements (*h1*, *h2*, …, *h7*: table 1). Based on self-organizing these objects in a non-parametrical toroidal 2-D space, the unsupervised clustering proceeds for $t = 1 \times 10^6$ time steps. To difficult our problem, and in order to have an idea on the behaviour of our algorithm in classifying different samples based on their features - since the numbers of samples were not elevated-, we decide to give as an input not 20 samples, but 60, cross-validating the final results. In order to do so, sample 21 is a copy of sample 1, sample 22 equals sample 2, …, sample 40 equals sample 20. Likewise, sample 41 is a copy of sample 1, …, sample 60 equals sample 20. These samples belong to two different classes (scallop and non-scallop). Scallop samples are represent by ID numbers 7, 12, 14, 16 and 18. As well as samples 27, 32, 34, 36, 38, 47, 52, 54, 56 and 58. All the remaining samples (out of 60) belong to the other class, i.e. non-scallop. After the unsupervised and self-organized clustering process - with 60 samples x 7 features each - is finished (see fig. 4 and 5), the first 20 samples are used as markers or reference points (filled square points in fig. 5), and via $k$-NNR nearest neighbour rule classification [10,13] the remaining 40 samples (non-filled square points in fig. 5) are classified (we used $k = 3$ neighbours; $k$ must be always an odd number). In order to do so, for each sample $i = 21, \ldots 60$, we computed their first $k = 3$ marker neighbours on this non-parametric toroidal 2-D space. The majority of those marker label values - considered for each still unclassified sample - give them the respective classification result.

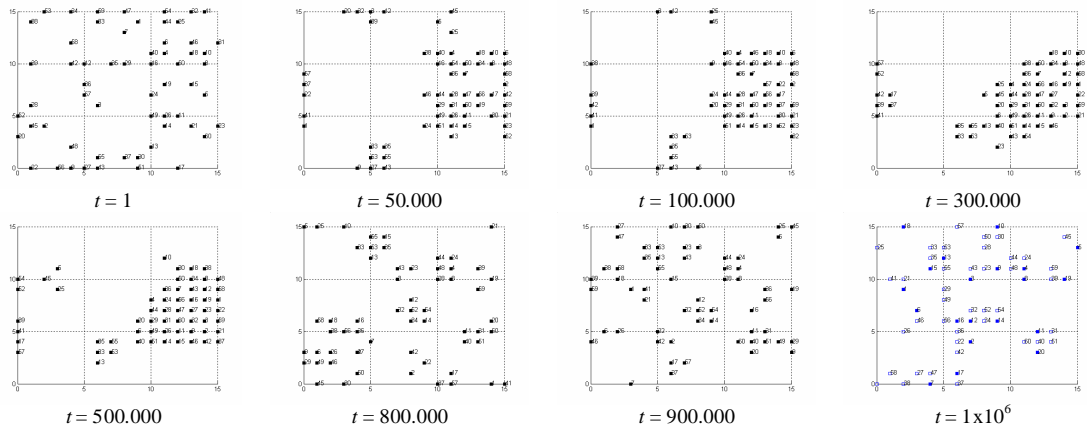

Fig. 4 – Ant-like clustering system on Shellfish Larvae Data. From $t=1$ to $t=1 \times 10^6$, a sequential clustering task of 60 data-objects (60 image samples each represented by 7 features) performed by 6 ants on a 15 x 15 non-parametric toroidal grid (processed in 57 seconds on a normal computer). At $t=1$, all items are randomly allocated into the grid. As time evolves, several homogenous clusters emerge due to the collective ant colony action, and as expected the global entropy decreases [37]. Final result could be seen in greater detail at fig. 5.

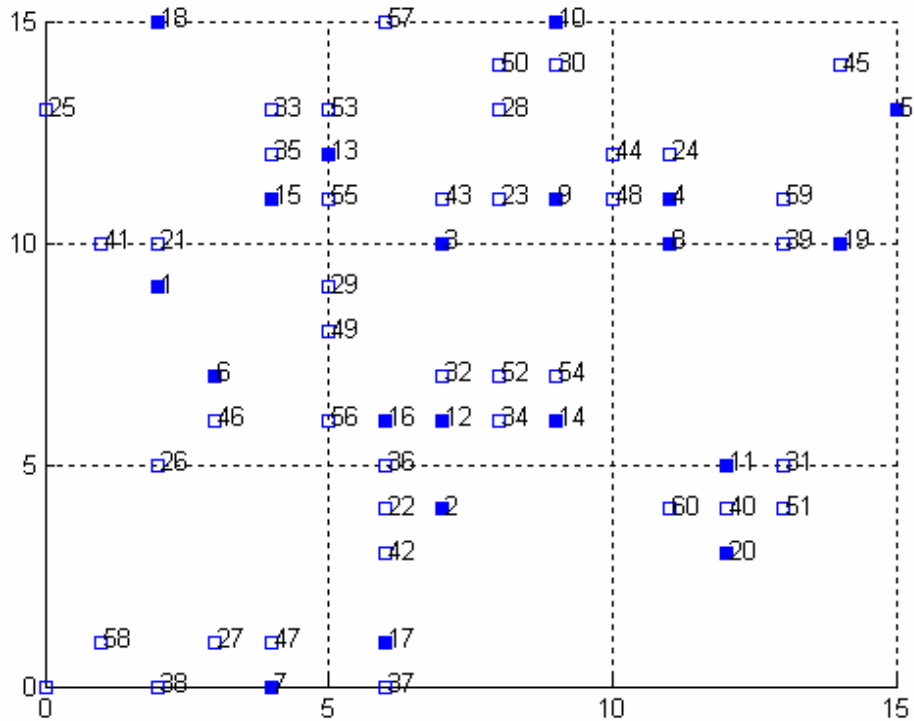

Fig. 5 – Ant-like clustering system on Shellfish Larvae Data. Final result at $t=1 \times 10^6$. Filled square items are now markers in order to classify non-filled square items via $k$-NNR ($k = 3$), on this toroidal grid.

Final results point out to not even 1 error out of 40 samples classified ($i = 21, \ldots 60$), conducting us to a successful recognition rate of 100%. It is interesting however to follow how sample ID 58 (bottom-left corner in fig. 5), a scallop shellfish larvae, was classified, since it was one of those at risk. Their first 3 nearest marker neighbours were ID samples 18 (scallop), 7 (scallop) and 17 (non-scallop). If for some stochastically dynamical reason, marker samples ID 18 or 7 were not there in the geographical proximity, marker sample ID 20 (non-scallop) should instead be considered and the final result would be different. Nevertheless, and for any of these cases, the final self-organized stigmergic map [37] achieved is highly robust. If for example, we use $k = 1$ in the finally *k*-NNR classification, which normally is not prudent, we also arrive at a full recognition rate of 100%.

## 6. Conclusions

We have described the use of invariant moment shape features for the discrimination of scallop larvae in microscope images. Images of a sample of scallop and non-scallop larvae covering a range of maturities have been analysed. Based on a limited data set, the distribution of selected features suggests the feasibility of automated identification. In order to scale up the automated identification system to more complex situations, an unsupervised self-organized clustering algorithm was introduced, followed by simple *k*-NNR nearest neighbour classification. The system performed well over 40 samples, and it shows promising possibilities to overcome new and different larvae samples. On the other hand, the present method shows how stigmergy can easily be made operational. Indeed, it is a promising first step to design groups of artificial agents which solve problems: replacing coordination (and possible some hierarchy) through direct communications by indirect interactions is appealing if one wishes to design simple agents and reduce communication among agents. Finally, stigmergy is often associated with flexibility: when the environment changes because of an external perturbation, the insects respond *appropriately* to that perturbation, as if it were a modification of the environment caused by the colony's activities. In other words, the colony can collectively respond to the perturbation with individuals exhibiting the same behavior. When it comes to artificial agents, this type of flexibility is priceless: it means that the agents can respond to a perturbation without being reprogrammed to deal with that particular instability. In our context, this means that no classifier re-training is needed for any new sets of data-item types (new classes) arriving to the system, as is necessary in many classical models, or even in some recent ones. Moreover, the data-items that were used for comparison purposes in early stages in the colony evolution in his exploration of the search-space, can now, along with new items, be re-arranged in more optimal ways. Classification and/or data retrieval remains the same, but the system re-organizes itself in order to deal with new classes, or even new sub-classes of larvae species. This task can be performed in real time, and in robust ways due to system's redundancy. Further work on the project will focus on: (a) collection of a more extensive data set; (b) implementation and evaluation of different classification techniques [13,21,2,43] including the present proposal; (c) development of a robust segmentation technique [8,17,23,26,27,44]. Results from preliminary investigations into automatic identification using classical computational pattern recognition methods applied to digitised microscope images have been published [3-7]. Subsequent work has identified two significant topics for further study: (i) image segmentation methods since some translucent larvae are difficult to separate from their background using purely density-based segmentation methods; (ii) further development of shape recognition methods, especially the recognition of translucent objects in the presence of occlusions. The major research outcomes will be advancement of two areas of pattern recognition: (i) segmentation including spatial model based methods; (ii) shape recognition methods, including occlusion handling. In addition, the work will make a significant advancement to the understanding of the applicability of machine vision methods to image analysis problems in marine biology and related areas.